\documentclass[]{interact}

\usepackage{epstopdf}
\usepackage[caption=false]{subfig}

\usepackage[numbers,sort&compress]{natbib}
\bibpunct[, ]{[}{]}{,}{n}{,}{,}
\renewcommand\bibfont{\fontsize{10}{12}\selectfont}
\makeatletter
\def\NAT@def@citea{\def\@citea{\NAT@separator}}
\makeatother

\theoremstyle{plain}

\usepackage{verbatim}
\usepackage{xcolor}
\usepackage{algorithm}
\usepackage{algpseudocode}

\theoremstyle{definition}

\theoremstyle{remark}

\newcommand{\hakan}[1]{{#1}}
\begin{document}

\articletype{FULL PAPER}

\title{Multi-step planning with learned effects of partial action executions}

\author{
\name{Hakan Aktas\textsuperscript{a}\thanks{CONTACT Hakan Aktas Email: hakan.aktas1@boun.edu.tr}, Utku Bozdogan\textsuperscript{a} and Emre Ugur\textsuperscript{a}}
\affil{\textsuperscript{a}Department of Computer Engineering, Bogazici University, Istanbul, Turkiye}
}

\maketitle

\begin{abstract} 
\\\resizebox{20pc}{!}{\includegraphics{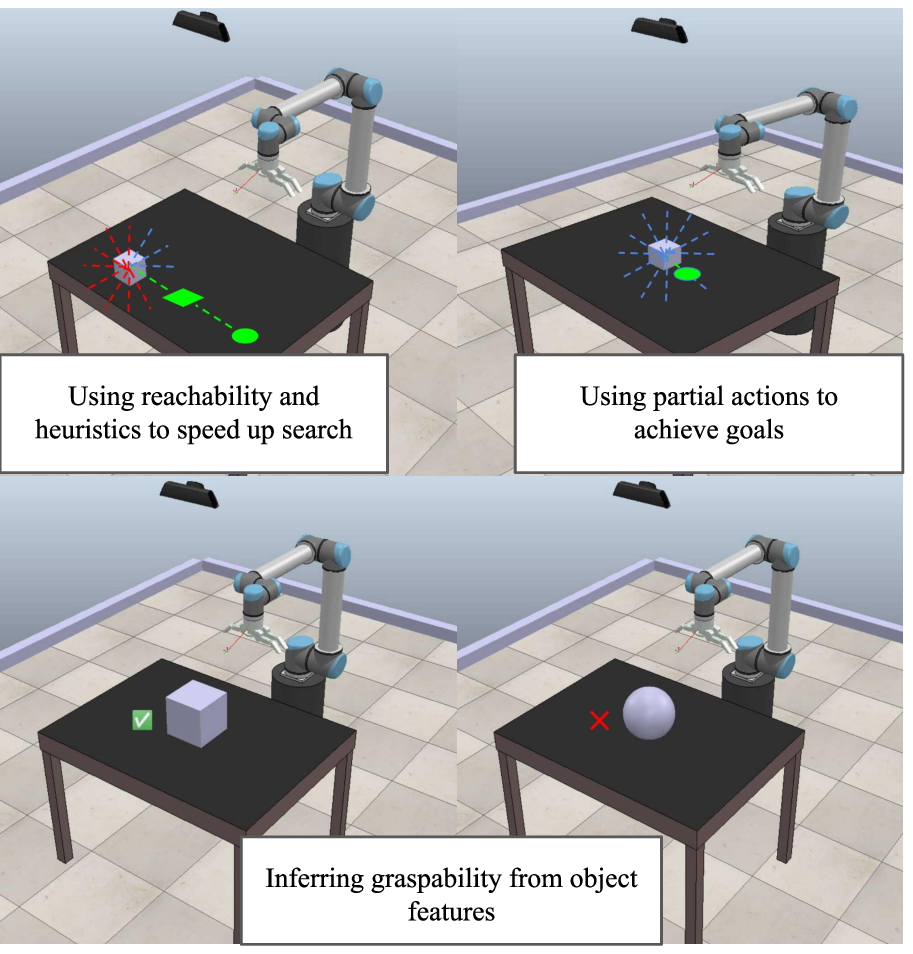}}\\
\hakan{In this paper, we propose a novel affordance model, which combines object, action, and effect information in the latent space of a predictive neural network architecture that is built on Conditional Neural Processes. Our model allows us to make predictions of intermediate effects expected to be obtained during action executions and make multi-step plans that include partial actions. We first compared the prediction capability of our model using an existing interaction data set and showed that it outperforms a recurrent neural network-based model in predicting the effects of lever-up actions. Next, we showed that our model can generate accurate effect predictions for other actions, such as push and grasp actions. Our system was shown to generate successful multi-step plans to bring objects to desired positions using the traditional A* search algorithm. Furthermore, we realized a continuous planning method and showed that the proposed system generated more accurate and effective plans with sequences of partial action executions compared to plans that only consider full action executions using both planning algorithms.}
\end{abstract}


\section{Introduction}

From a robotics standpoint, learning to predict the effects of a robot's actions beforehand would be a beneficial skill for robots, since this can prevent potential failures and dangerous situations for the robot and those around it, and enable planning for achieving certain goals.
Planning for multi-step tasks in the real world is difficult, and a generalized approach to solving this problem is even more so. Due to its difficulty, previous works compromise on certain aspects, such as predefining effect categories or object categories, simplifying the task. Discretizing information in the continuous sensorimotor space with the purpose of high-level symbolic planning with symbols may result in inaccurate plans.\hakan{The current study is different from the previous studies in the sense that the previous studies only considered the final effect (and final effect category, and for this, they need to make clustering in effect space), whereas our study utilizes effect prediction trajectory along the action execution. Our study can sample any point along the predicted effect trajectory and use the predicted effect in planning. Additionally, a new motion planning algorithm is proposed as a continuous solution.}

In robotics, learned affordances have been used to choose objects for manipulation, to make discrete and continuous effect predictions given objects and actions, and to make  plans to achieve goals \cite{Jamone2016,Yamanobe2017,Zech2017,Taniguchi2018}. \hakan{However, to the best of our knowledge, there is no single framework that can achieve all the following,
\begin{itemize}
    \item predict the effects given objects and action
    \item predict required actions to achieve desired goals
    \item predict the movement trajectory of the objects in response to parametric actions
    \item make plans composed of a sequence of actions, including partial action executions, in order to achieve given goals
\end{itemize}
}

The contribution of our work is as follows:
\begin{itemize}
    \item A novel latent representation for affordances: Considering affordances as relations between objects, actions, and effects, our neural network-based architecture forms a representation that encodes object, continuous action, and continuous effect information in a single latent layer. 
    \item Accurate continuous effect prediction: Our system can predict the motion trajectories of the objects expected to be generated by parametric robot actions. We showed that the high prediction performance compared to a strong baseline also makes affordance-based continuous effect prediction.
    \item Multi-step planning with partial actions: We exploited the outcome prediction capability of our system for partial actions in order to form multi-step plans that may include full or partial action executions. \hakan{To show this, we both used a traditional planning method namely A* and we also proposed a new continuous adaptive planning algorithm that better utilizes our method.}
    
\end{itemize}

\section{Related work}


\paragraph*{Early affordances work on effect predictions}
In early works such as \cite{fitzpatrick2003learning}, objects were required to be recognized first, and object-specific affordances were learned from the robot's interaction experience with those objects, therefore the learned affordances could not be generalized to different/novel objects.  \cite{ugur2007learning} studies learning of traversability affordance where the LIDAR input was directly processed without any object detection step. Therefore, the `direct perception' aspect of affordance perception was realized in that work, however only a single pre-defined affordance category was addressed. Effect categories and affordances were discovered by the robot in \cite{ugur2011going} via unsupervised clustering techniques. However, unsupervised clustering results depend on the particular clustering method and the feature space that might be composed of values with different metrics such as distance, frequency, angle, etc. In \cite{ugur2011goal}, this is taken one step further and hierarchical clustering is made over channels for better effect category discovery. Both \cite{ugur2011going} and \cite{ugur2011goal} also enable forward chaining for multi-step planning. 
Bayesian Networks were used \cite{montesano2007modeling, montesano2008learning}, enabling bidirectional predictive capabilities using the robot's own interaction experience, but clustering is performed on object features and effects in a predefined manner. In this paper, we also study acquiring bi-directional prediction capabilities. Different from previous work, we do not find effect clusters or do not only aim to predict the final effect. Instead, our system aims to learn how to predict the complete effect/action trajectory during action execution. 

\paragraph*{Learning affordances for planning}
\cite{ugur2015bottom} learned affordances for symbolic planning, by again clustering effect categories and using object categories as a collection of effect categories obtainable by actions available to the robot. This enables the representation of nonlinear relations in planning, however, their discrete representation while making planning easy, makes the estimations approximate, increasing long-horizon planning errors. This work was validated on a real-world setup with simple actions such as poke, grasp, and release, and also with a more complex action which is stack. The experience of the robot enabled it to gain experience from the simple actions and after experiencing the stack action enabled it to generate a valid plan for the stacking action. This framework was extended in \cite{ugur2015refining} by enabling the robot to progressively update the previously learned concepts and rules in order to better deal with novel situations that appear during multi-step action executions. Similar planning capabilities were obtained by the robot using deep encoder-decoder neural networks with binary bottleneck layers \cite{ahmetoglujair} and with multi-head attention mechanism \cite{ahmetogluhum} by directly processing the pixel images instead of using hand-coded features.
In \cite{ames2018learning}, probabilistic planning with symbols using parameterized actions was applied to a real robot task, showing that continuous tasks can be performed with discrete planners using parameterized behaviors. Different from the previous work where only final outcomes were used for planning, our system can exploit intermediate effects expected to be observed for example in the middle of the action execution for planning.

\paragraph*{Learning visual grasp affordances}
Using RGB/RGBD images for predicting affordance classes or pixel-wise affordance labels for object manipulation has become popular in recent years \cite{nguyen2016detecting, do2018affordancenet, mi2019object, hamalainen2019affordance, chu2019toward, thermos2021joint} and was shown to be a feasible approach for learning how to grasp different objects. Similarly, \cite{zhang2022inpaint2learn} are able to learn the affordances of objects such that they can place objects/humans in correct poses in a scene and also choose the correct object type to place in a given scene.\hakan{ \cite{saito2021tool} and \cite{mar2017can} extracted not only object features but also the visual features of tools from their images to generate motion trajectories using tools as a part of affordance.}\cite{ruiz2020geometric} uses point clouds to learn general geometric features from object interactions, enabling them to place objects in a scene correctly. In \cite{khazatsky2021can}, a generative model learned from interaction images was used to propose potential affordances. The aim was to learn a generalizable prior from interaction data and then utilize it to propose reasonable goals for unseen objects. These goals were then attempted to be executed by an offline RL policy, learned from interaction data, and tuned online efficiently to adapt to unseen objects. However, continuous effect prediction and multi-step planning aspects were not  addressed in these studies.

\paragraph*{Affordances for efficient learning}
Affordances can also be used to reduce the search space in order to efficiently generate plans to solve long-horizon tasks. Recent approaches utilizing this idea extended the definition of affordances to represent not only single-step action-effect knowledge but as action feasibilities \cite{xu2020deep}, or intents that are similar to goals \cite{khetarpal2020can} in order to make affordances useful for multi-step plans. However, the feasibility concept of \cite{xu2020deep} accepts a grasp action achieving nothing as afforded, and would also accept a grasp action as afforded regardless of whether it was an appropriate grasp for an object. While \cite{khetarpal2020can} overcame this with intent representation, intents were specified a priori in their work and although a sub-goal discovery direction was proposed for learning, it was not explored.

\paragraph*{Learning continuous effect trajectories}
Similar to our work, \cite{seker2019deep} and \cite{tekden2020} learned to predict the full motion trajectory of an object, using the robot's own interaction experience, and top-down images of the objects. These studies are important for their inclusion of the temporal aspect of effects. The utility of different features was also investigated, such as hand-crafted shape features, CNN-extracted features, or support point features extracted from a neural network. The authors used these features and the interaction experience to train recurrent neural networks and were able to accurately predict trajectories resulting from a lever-up action in a real-world setting with multiple objects. A common shortcoming of the aforementioned methods which make predictions for action or effects is related to the use of recurrent methods for long-horizon tasks. The use of recurrent networks such as LSTMs \cite{hochreiter1997long} or GRUs \cite{cho2014learning} is shown to be effective for short-term  predictions. 
In this paper, we compare the effect prediction capabilities of recurrent neural network-based systems and our conditional neural process-based system.

\section{Proposed Method}

\subsection{General Architecture}

\begin{figure}[t]
    \centering
		\includegraphics[width=1\textwidth]{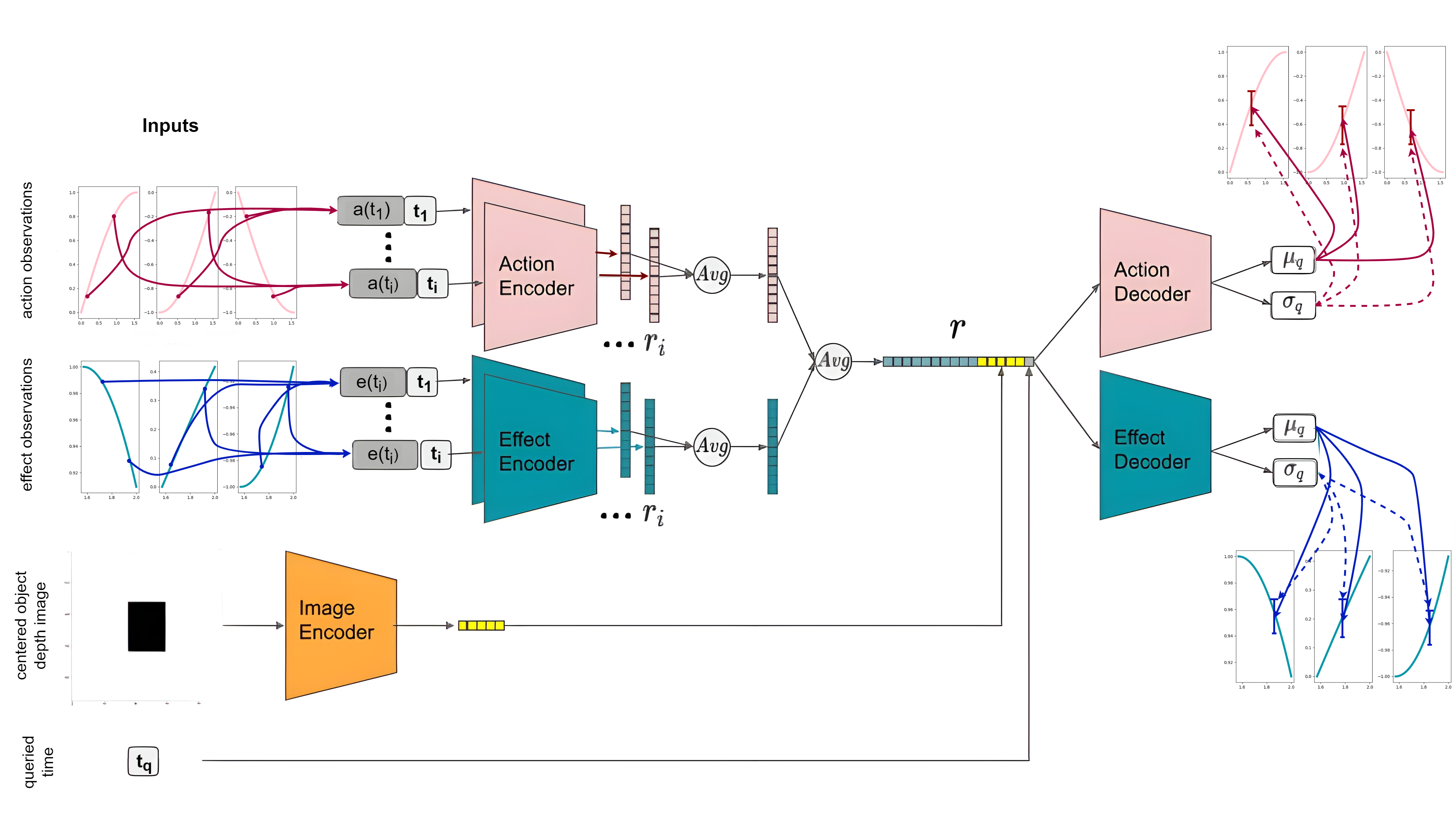}
		\caption{An overview of the proposed model. Given object image and action or effect information at any time-point, our system can generate the effect trajectory and action execution trajectory.}
		\label{fig:Figure_1}
\end{figure}

We propose a framework that can learn (i) predicting the effect trajectory given the initial object image and action execution trajectory, and (ii) finding the required action execution trajectory to achieve a desired effect on a given object. Our system is built on top of Conditional Neural Processes (CNPs) \cite{garnelo2018conditional} that bring together the inference potential of Gaussian Processes and the training of neural networks with gradient descent by learning a prior from data.  As we would like to predict both actions and effects from objects and (possibly missing) actions and effects, we propose a neural network structure that takes the object image as input together with action and/or effect values at different time points. In other words, given the initial object image, our network can be conditioned with action and/or effect values at different time points in order to generate action and effect values for all time points during action execution. The general structure of the proposed system is shown in Fig.~\ref{fig:Figure_1}. As shown the system is composed of two encoders that process action and effect information at different time points, an image encoder that processes the depth image of the object, averaging operations to acquire a common object-action-effect representation (r), which in turn can be used to predict action and effect values at other time points using the corresponding two decoders.

In detail, inspired by Deep Modality Blending Networks \cite{seker2022imitation}, our system encodes information coming from different channels (object, action, and effect) into a common latent space, using a weighted average of encoded modality representations, facilitating information sharing, and also providing a regularization effect on the representations learned, similar to dropout \cite{srivastava2014dropout}.

In our implementation, the system is object-centric. An action is defined in terms of the distance of the robot's end-effector to the \hakan{object's initial position and also the orientation of the gripper along with a binary value that shows whether the gripper is open or not}. An effect is defined as the displacement \hakan{and the change in orientation} of the object from its starting position throughout an action. A depth image of the object is included as an external parameter $\gamma$ for the action. \hakan{$a_t$ and $e_t$, represent the action and the effect at time $t$ respectively and since the object is independent of the $t$, it is represented by $o$ and is equal among the interaction trajectory $D_d$.}
\begin{equation}
    D_{d} = (\{a_{t},e_{t},o,t\}_{t=0}^{t=1})_{d}
\end{equation}
is an interaction trajectory where $1 \leq d \leq m$ and  $m$ is the number of trajectories in the data set $D$.  $0 \leq t \leq 1$ is a phase variable  in control of the passage of time where $t \in \mathbf{R}$. \hakan{Observation points sampled from these interaction trajectories are used as input to the corresponding channels of the model.} Each channel is encoded separately by its own encoder and the latent representations are subjected to a weighted averaging operation:

\begin{equation}
    r_{i} = \sum h_{a}(a_{t_{obs_{i}}}) * w_{1} +  h_{e}(e_{t_{obs{i}}}) * w_{2}
    \label{eq:affordx1}
\end{equation}

where $w_{1} + w_{2} = 1$, \hakan{ $h_a$ and $h_e$  represents action encoder and effect encoder respectively, and $a_{t_{obs_{i}}}$ and $e_{t_{obs_{i}}}$ represent the action and the effect observation points sampled from observation trajectory $i$. Each latent representation constructed by encoders is represented by $r_i$  and at each iteration $k$ latent representations are constructed. $k$ is selected randomly at each iteration. The $r_i$ values obtained from these calculations are then averaged:}  

\begin{equation}
    r = r_{1} \oplus r_{2} \oplus r_{3} \oplus ... \oplus r_{k}
    \label{eq:cnp2}
\end{equation}

where the commutative operation used is averaging. Then, depth image features and target time-step are concatenated to form the merged representation $r_{mrg} = \{r,f(\gamma_{o}),t_{target}\}$ \hakan{ where $f(\gamma_{o})$ represents the latent representation obtained by passing the object's depth image through the image encoder and $t_{target}$ represents the queried time (shown as $t_q$ in the Figure \ref{fig:Figure_1}).}

This merged representation is then decoded by separate decoders, each corresponding to a different channel. This merged representation is decoded at the action decoder \hakan{($g_{a}$)} by
\begin{equation}
    g_{a}(r_{mrg}) = (\mu_{a_{t_{target}}}, \sigma_{a_{t_{target}}}),
\end{equation}
and the effect decoder \hakan{($g_{e}$)} by
\begin{equation}
    g_{e}(r_{mrg}) = (\mu_{e_{t_{target}}}, \sigma_{e_{t_{target}}}),
\end{equation}
to yield predictions for action and/or effect for the target time step shown in Fig~\ref{fig:Figure_1}. \hakan{Here, mean values are used as predictions, and variances are only used to calculate the loss of the system and can also be used to determine how confident the system is with the current predicted value.}

The latent representation (r) can be viewed as the shared affordance representation for actions $a$ and effects $e$ for different objects $o$. Learned affordances can then be used to predict the effects of actions or the required action to generate target effects. This model can be used to create multi-step action plans to achieve goals beyond single-action executions by chaining the predictions.

\subsection{Training}

At each training iteration, $k$ observations are sampled uniformly at random from a randomly selected interaction trajectory $D_{d}$ where $1 \leq k \leq obs_{max}$, $k \in \mathbf{N}$ is also a uniform random number, and $obs_{max}$ is a hyper-parameter denoting the maximum number of observations the model is allowed to use during one iteration. These observations are then encoded and aggregated. A cropped object depth image is encoded separately on a CNN encoder network, and the resulting vector is concatenated at the end of this aggregated representation. Before a prediction can be made, a target time step is also concatenated to the image features. Finally, this merged representation is decoded to yield predictions for action and/or effect for the target time step shown in Fig~\ref{fig:Figure_1}. Gradient descent is used with the loss function (\ref{eq:cnmp_loss}) with Adam optimizer \cite{kingma2014adam}.

\begin{equation}
    \mathcal{L}(\theta, \delta) = - log P(y_{j} | \mu_{j}, softmax(\sigma_{j}))
 \label{eq:cnmp_loss}
\end{equation}

After training, the network is able to predict the entire interaction trajectory given \hakan{any novel point(s) on a trajectory}. An A* planner \cite{Hart1968} is then used on top of the network to solve tasks requiring multiple actions and steps.

\begin{figure}[t]
    \centering
		\includegraphics[width=0.6\textwidth]{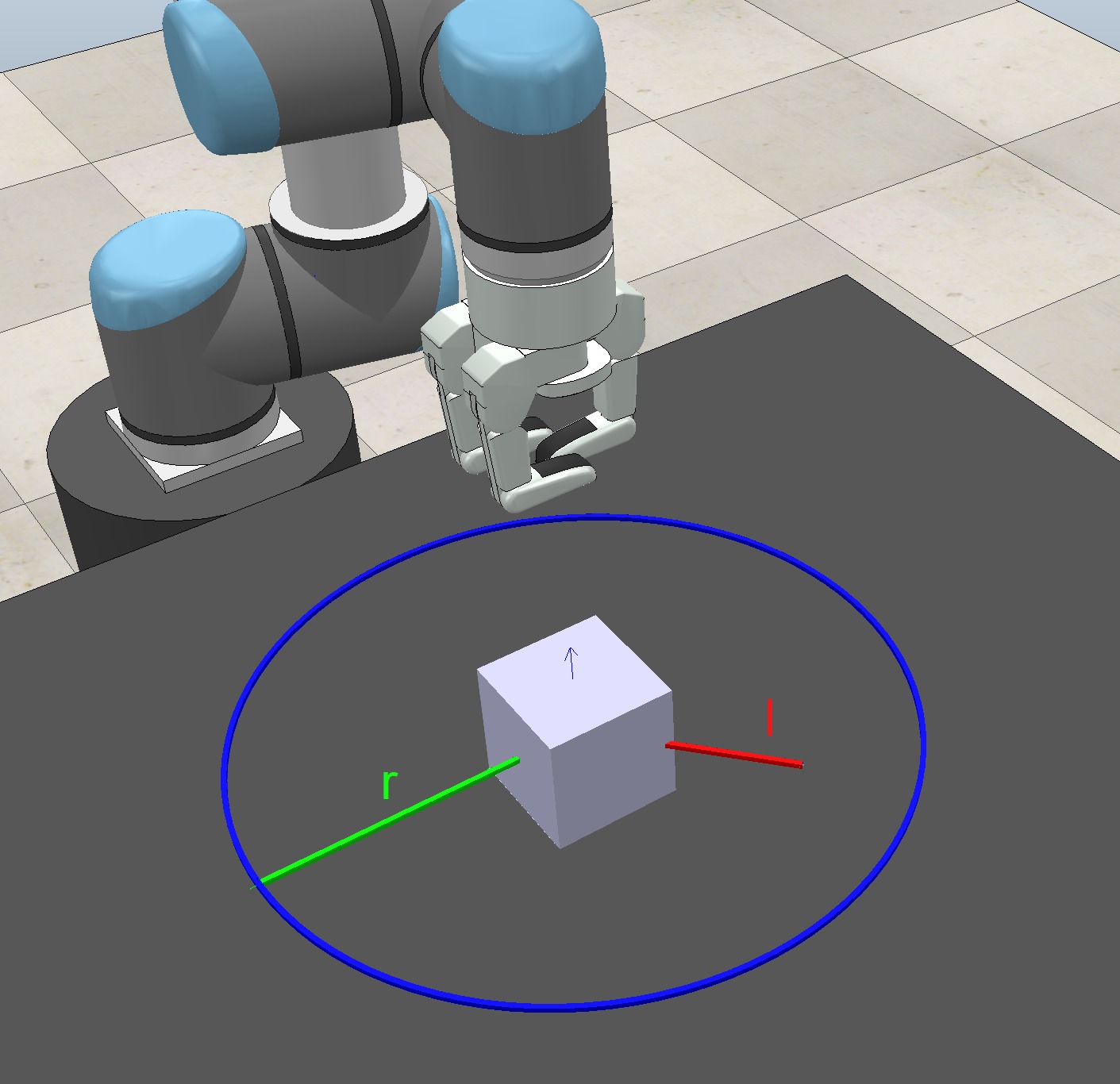}
		\caption{Scene showing the parameters of a push action around an object.}
		\label{fig:Figure_2}
\end{figure}

\subsection{Actions}

Our arm-hand robot is equipped with \hakan{three actions, namely push, grasp, and rotate}. The parametric push action is specified by an angle $\theta \in [0,2\pi]$; a push distance $l = 0.05$ and a $radius = 0.2$ both in meters. The gripper starts the push execution from the red circumference of a circle of radius ($radius=20cm$) centered around the object (see Figure~\ref{fig:Figure_2}), at an angle $\theta$ and pushes the object $l$ meters from its center of mass. 

\begin{figure}[t]
    \centering
		\includegraphics[width=0.7\textwidth]{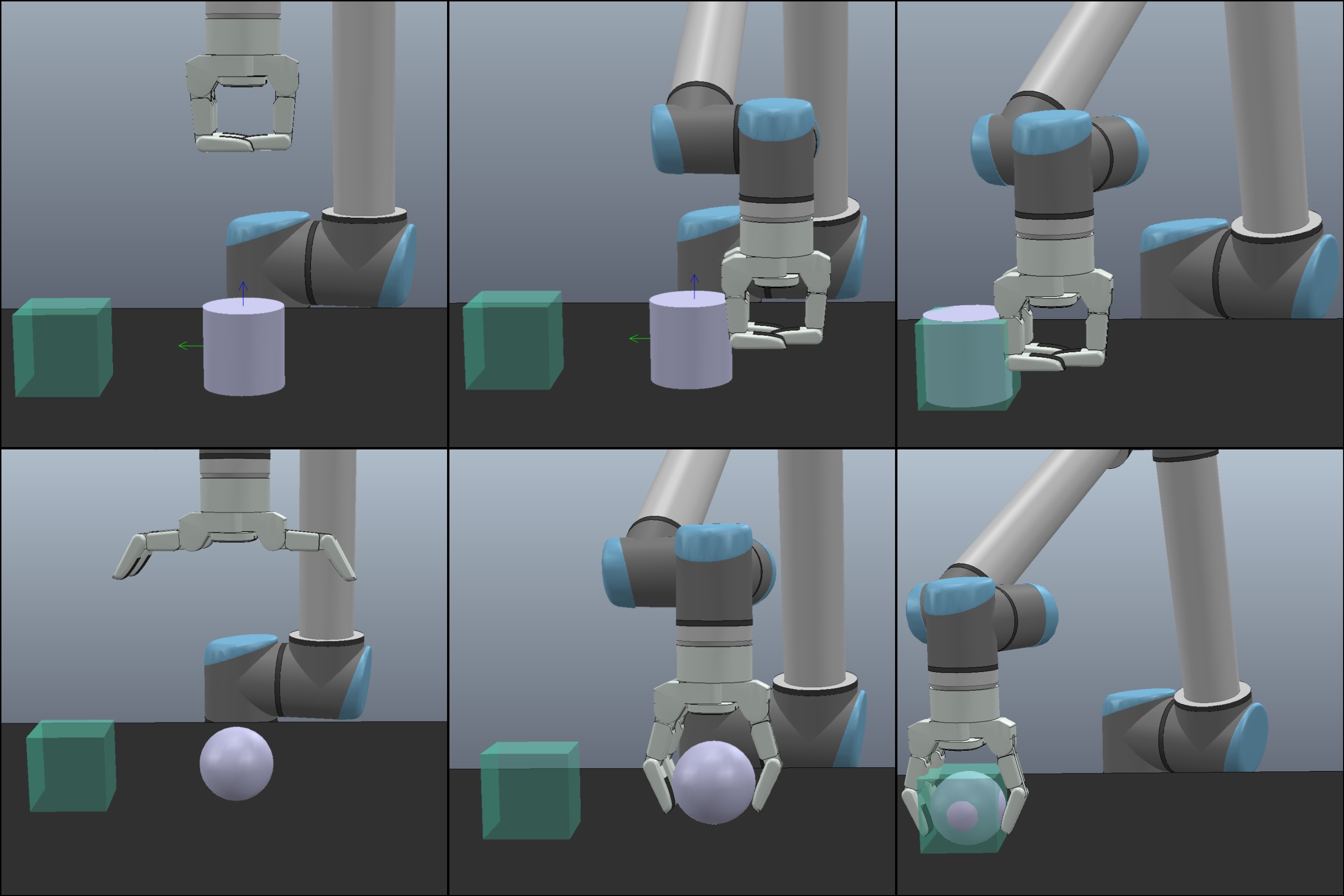}
		\caption{The actions used to push a non-rollable object (cylinder in the figure) and a rollable object (sphere in the figure). All push actions used to train the network are composed of two steps as shown: approach the object and bring it to the desired location. The desired location is shown with the green shaded area. Based on the angle to push the orientation of the gripper changes. Note that, the sphere is not elevated during this action.}
		\label{fig:push-figure}
\end{figure}

Larger-sized objects may be displaced more as a result of this setup. The model is expected to learn the rollability and pushability affordances from these interactions and based on the object shape be able to predict the trajectories of  rollable objects such as spheres or lying cylinders and non-rollable objects such as cuboids or upright cylinders.\hakan{Since the rollable objects are unstable if the push actions used for non-rollable objects are used for the rollable ones, the objects  keep moving even after the action is completed which makes it impossible to make multiple step plans using these objects. To overcome this challenge we devised a new action that keeps the rollable objects from rolling off after the action is completed. Using this action, our model can make different plans for objects with different behaviors to achieve the same goal which is to take the object to the desired object location. This action is executed by lowering the open gripper on top of the object, closing it to keep the object in place, and then taking it to the desired location. Although this action is a little different than a usual push action, we will call it push for simplicity. The difference between the action for a not rollable and a rollable object can be seen in Figure \ref{fig:push-figure}.}

Grasp actions, on the other hand, are realized by lowering the open gripper to grasp position, and attempting to grasp an object by closing the gripper and lifting the gripper up. Our model is expected to learn the graspability affordance from these interactions. Based on object size and shape, it should also be able to predict the interaction trajectories.  \hakan{Rotate actions are similar, and realized by lowering the open gripper on top of the object, closing the gripper to keep it in place, and rotating the gripper to rotate the object.}

\subsection{Planning with partial action executions}

\hakan{For the planning with partial action executions, we experimented using two approaches. We used A* planning algorithm to show that our model can be used with traditional planning methods. We also proposed a proprietary planning algorithm that utilizes the full potential of our method, and compared its results with the A*.}

The A* planner (with Euclidean distance to the goal heuristic) is used to generate a sequence of actions to move the object from its initial position to a goal position. Each branch in the search tree corresponds to either a push action or a grasp action. \hakan{As grasp and rotate actions are not parameterized, there are single branches for grasp and rotate actions}. On the other hand, as a push action might be applied from different approach angles and for different push distances, the range of possible approach angles and push distances are discretized and used to create multiple branches from the same node in the search tree. The initial 20\%, 40\%, 60\%, 80\%, and 100\% segments of push action were considered during the search. Additionally, the push direction was discretized into 36 directions. Therefore, the branching factor was set to 180.

The planner uses predictions of actions with different parameters to update the predicted location of the object. \hakan{The actions are assumed to be known and the effects of those actions are generated by the proposed deep neural network model. To generate a sequence of actions, the effect predictions that are generated using the segmented actions are traversed in the search tree to find a sequence with a desirable outcome.} The search is completed if the difference between the predicted and goal object position is less than 2 cm. Our model is able to work with continuous inputs, however, the planner can only propose a finite amount of actions due to the mentioned discretization design choice. 

Interactions with single and multiple actions are generated, which also include partial actions. Partial actions are when an action is started to be applied. However, it is not completed, i.e. it is executed partially. For example in a push action, the push may be cut short before the gripper even contacts the object, or the gripper reaches the center of mass of the object, meaning that the object has already started being pushed, but the push is not completed yet. Importantly, our model is trained only with full-action interactions. Yet, our model can generate the effect at any desired time point, and therefore it can predict the consequence of partial action executions. The planner can use the effects of such partial action executions to generate plans with finer resolutions (compared to plans that can only include full action executions).

\hakan{We also propose a continuous method for motion planning to take full advantage of our model's capabilities. In our method, we realize the push actions in two steps. First, the direction to the desired location is found, and then the object is pushed to the desired location either partially or fully. To find the direction in which the object needs to be pushed, we uniformly sample a random point along the circumference of the circle the push actions are defined on (section 3.3). Then this point is given as input to the effect channel of our model. The input for the action channel can be set to any value since we set the weight of the action channel to zero and the weight of the effect channel to one during this process. Assuming the time is normalized, we set $t_{target}=1$ since the points along the circumference of the circle correspond to the last points of the trajectories that lead to that point. After providing the object depth image as input to the model, we obtain a prediction in the output of the effect decoder which is close to the point we set as input to the effect encoder. }

\hakan{Our aim is to make this predicted point as close to the desired location as possible since the closest point to the desired location will be along the direction to the desired object location. To this end, we calculate the mean squared error between the predicted point and the desired final object location. Using the value obtained from this calculation as the loss, we apply gradient descent to the input of the effect encoder to adjust the input until we get the input that produces the closest point to the desired location. Note that the rest of the model is not trained during these calculations. Gradient descent is only applied to the input of the effect encoder to adjust it, not to train any part of the model. Using this, our system is able to find the direction to the desired object location after sufficient iterations are carried out. This way the direction of the action is found in a continuous manner.}

\hakan{After finding the direction, the system is queried using $t_{target}$ values between $0$ and $1$ until the desired object location is achieved. If the desired location is in between the initial position and the range of the full push action, the system is queried until the predicted object location is not getting closer to the desired location and thus a partial action is found. If the desired location is on the end of the action range or beyond it (another action is needed), then the system is queried until $t_{target} = 1$ . This way, any segment of any action that starts from the starting point of the action can be used as a partial action trajectory during execution. If more than one action is needed, the predicted object position at $t_{target} = 1$ is used as the object starting point of the next action in the action sequence. This process is repeated before execution until the desired object location is reached. }  

\hakan{Using this method, the sequence of actions needed to bring the object to the desired location in the 2D plane is determined before the execution starts. After each execution, the algorithm also checks whether the object is in the planned location and whether any goal change has occurred to adapt to any unexpected changes. After each execution, if the object is not in the expected location or the goal has changed in between executions, a new plan is made to bring the object to the desired location, and the initial plan is discarded. If the object used has different rollability affordance characteristics depending on which of its faces it is standing on (like the cylinder), and if the object falls over during action execution (its rollability affordance is changed), the planner revises the plan and makes a plan that is suited for the other affordance characteristic of the object by using the new top-down depth image of the object. The grasp and rotate actions can be chained to the end of the planning in the xy plane if the object is wanted to be in a certain height or orientation. The desired height or rotation can be achieved by querying until the desired height or rotation is obtained using the grasp and the rotate actions respectively.}

Note that an action can be applied only if the object is reachable by the robot. Our model is expected to learn the reachability affordance from interactions  experiences and actions are only considered during planning if the object is reachable.

\section{Experiments results}

\subsection{Effect prediction performance}

\begin{figure}[t]
  \centering
  \subfloat[\label{fig:shapeandmodel}]{\includegraphics[width=0.3\linewidth]{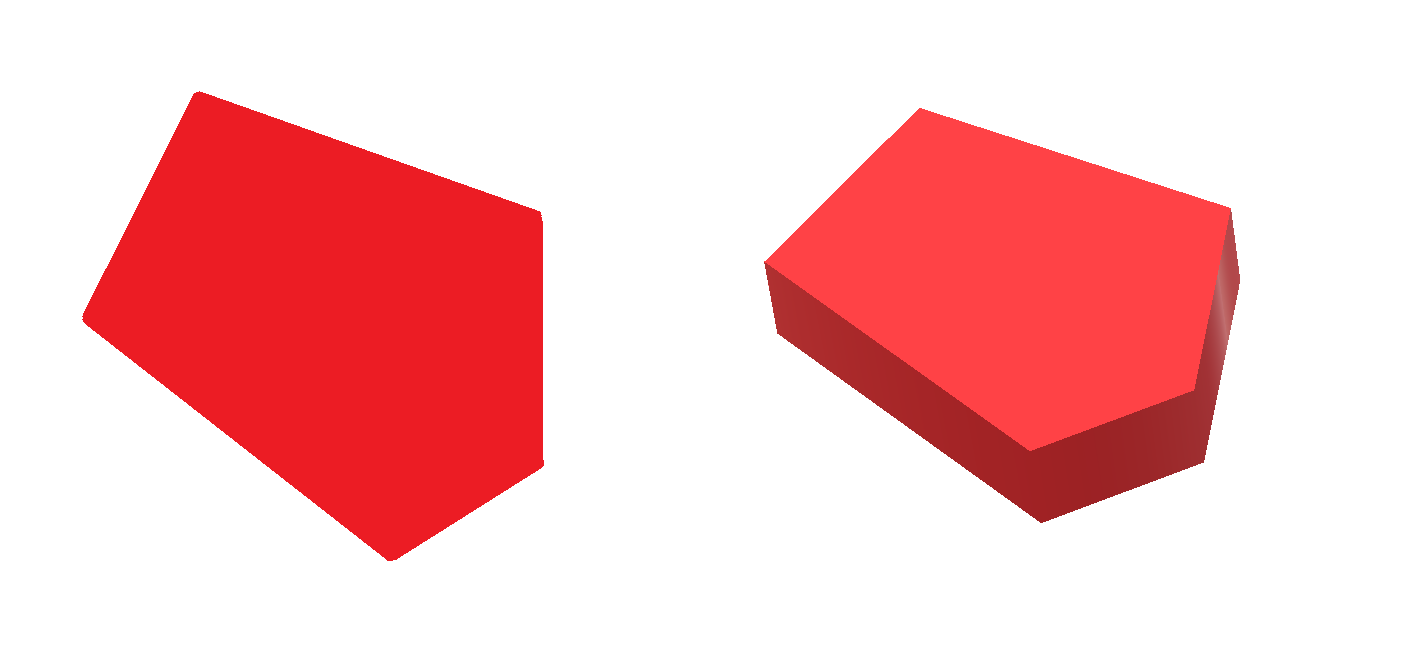}}
  \subfloat[\label{fig:shape82}]{\includegraphics[width=0.6\linewidth]{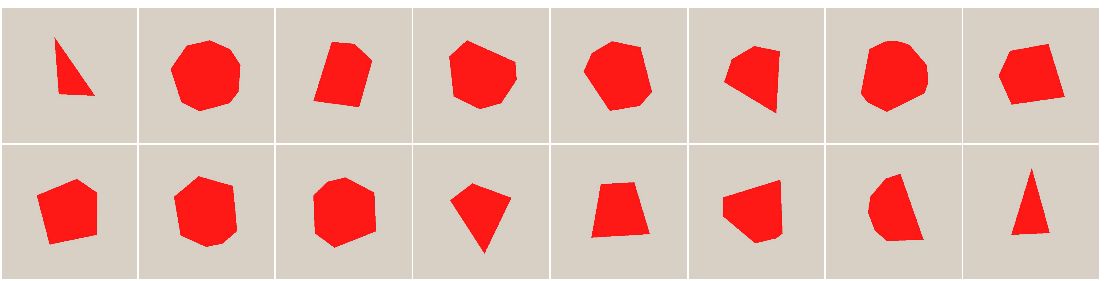}}
  \caption{(a) A sample random shape (pentagon) and the corresponding object. (b) A number of randomly generated sample shapes. Reused with permission from \cite{seker2019deep}.}
  \label{fig:shapes}
\end{figure}

\begin{figure}[b]
	\begin{center}
		\includegraphics[width=0.95\textwidth]{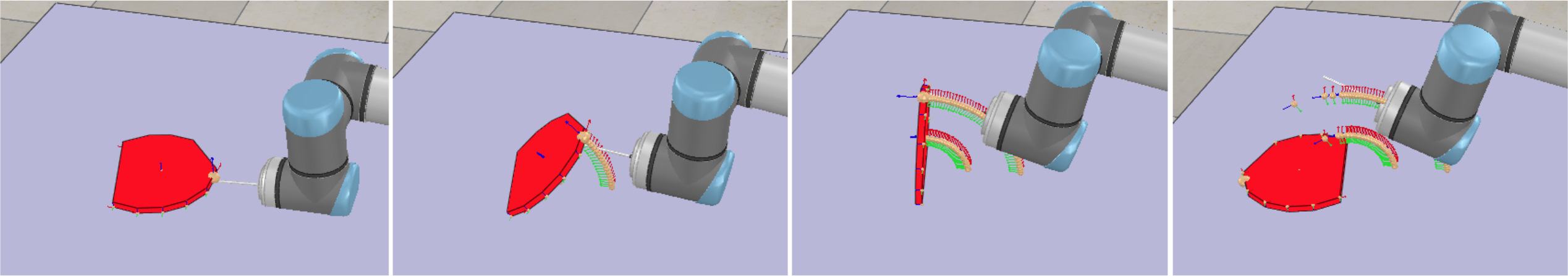}
		\caption{Example lever-up action in the simulator, reused with permission from \cite{seker2019deep}.}
		\label{fig:Figure_3}
	\end{center}
\end{figure}

In this paper, we propose a novel system to predict the effects given objects and actions. Furthermore, our system can generate complete motion trajectories of objects as effects rather than their final positions. In order to assess the performance of our system, we compare our system with a recent study that can also predict motion trajectories of objects using CNN and Long short Term Memory (LSTM) model \cite{seker2019deep}. We used the same dataset that \cite{seker2019deep} used where lever-up actions were applied to objects with different geometric shapes from different points. An example lever-up action is shown in Figure~\ref{fig:Figure_3}. The objects had different number of edges (between 3 and 8), sizes, and orientations. \hakan{Some examples of the objects used in this experiment can be seen in Figure \ref{fig:shapes}}. The objects were levered up from different contact points. The authors translated and rotated the top-down 128x128 grayscale image according to the contact point and lever-up direction in order to simplify the prediction problem. The dataset was separated randomly into 80\% training, 10\% validation, and 10\% testing as in \cite{seker2019deep}. 

The experiments were performed with 5-fold cross-validation and early stopping with one million iterations. For comparison, we gathered n-step predictions, taking 15 previous steps as observations, and compared the results with the ones reported in the same manner in \cite{seker2019deep}. As shown in Table ~\ref{table:Table_1}, the error between the predicted and actual positions of the objects in \cite{seker2019deep} varies between $0.90-1.00$cm, whereas the error was in the range of $0.3-0.4$cm in our system. \hakan{The total movement of the objects used is around 10 centimeters.} This comparison shows our model yields significantly lower error rates compared to a recurrent method since the output is predicted directly avoiding error accumulation for multi-step predictions.


\begin{table}
\tbl{N-step prediction errors obtained in our model and the LSTM model.}
{\begin{tabular}{|c|c|c|} \hline
\textbf{N-step}&  \textbf{Our Model}& \textbf{LSTM Model}\\\hline
\textbf{1-step (cm)} &  $0.340 \pm 0.044$  & $1.02 \pm 0.01$\\\hline
\textbf{2-step (cm)} & $0.352 \pm 0.046$  & $1.00 \pm 0.02$\\\hline
\textbf{3-step (cm)} &  $0.364 \pm 0.049$  & $0.96 \pm 0.03$\\\hline
\textbf{4-step (cm)} &  $0.375 \pm 0.051$  & $1.01 \pm 0.04$\\\hline
\textbf{5-step (cm)} &  $0.387 \pm 0.054$  & $0.91\pm 0.03$\\\hline
\end{tabular}}
\label{table:Table_1}
\end{table}


\subsection{Training Environment}
A simulated scene was constructed in CoppeliaSim \cite{coppeliaSim} as shown in Figure~\ref{fig:Figure_2}. A UR10 robot interacts with objects of different shapes and sizes by applying push and grasp actions on a tabletop setting. A Kinect sensor is placed above the table vertically such that the entire table is visible. The parts of the interaction where the object is potentially going to be displaced are recorded. The recorded information consists of action and effect data from once every 3 simulation steps (a single step is 50ms), which is chosen empirically, and a single depth image of the table with the object on top taken at the beginning of each interaction. The simulation dataset is split into training (\%80), validation (\%10) and test (\%10) sets.

\subsection{Single-action push and grasp effect prediction}
Different models were trained for push and grasp actions separately. The simulation data sets were always split between 80\% training, 10\% validation, and 10\% test data. For all the results reported in this work, 10-fold cross-validation was applied unless otherwise specified. The models were trained for one million iterations, without batches due to variable length of inputs, and early stopping was employed. The learning rate was set to $1\mathrm{e}{-4}$. All errors reported in meters denote the distances to a specified goal position. The predictions for a single action are fixed to take 25 time steps. 

For the push action, a data set made up of 500 trajectories was used. For each interaction, objects were chosen randomly and placed at the center of the table. An angle for the push $\theta \in [0,2\pi]$ was chosen randomly. The robot performed a complete push action and the resulting interaction data was recorded. The error in predicting the final position of the object was found to be around $0.02$m as shown in Table \ref{table:Table_3}. Similar to our analysis in the previous subsection, our system was shown to be effective in predicting object motion trajectories in push-like actions.

For the grasp action, a data set made up of 100 trajectories was used. Objects of varying sizes were randomly chosen to be placed at the center of the table. The robot then performed the grasp action and the resulting interaction data was recorded. The results for the grasp action are provided in Table \ref{table:Table_2}. In interpreting the performance and success of graspability prediction, if the change in height is larger than 0.1 meters, the grasp was assumed to be a success. If the change in height is less than 0.1 meters, it is a failed grasp. If the test data does not have a significant change in its z-axis coordinates but the predictions do, then this is a false positive and finally, if the test data has a significant change in z-axis coordinates but the predictions do not, then it is a false negative.  We have found that the robot had more difficulty grasping \hakan{rollable} objects of equal size, most likely due to the fact that a cube and an upright cylinder are both grasped by straight surfaces whereas large spheres and sideways cylinders are grasped by curved surfaces from points that are located above their center of mass, causing them to slip easier. \hakan{This difference can be seen in Table \ref{table:Table_2}}

\begin{table}
\tbl{Single grasp action prediction results on variable-sized objects of different rollability placed on a fixed location.}
{\begin{tabular}{|c|c|c|c|c|c|} \hline
\textbf{Object Type}&\textbf{Error (m)}& \textbf{True}& \textbf{True}& \textbf{False}& \textbf{False} \\
& & \textbf{Positive (\%)} & \textbf{Negative (\%)} & \textbf{Positive (\%)} & \textbf{Negative (\%)} \\\hline
\textbf{Rollable}&\textbf{$0.277 \pm 0.0086$}& 71.72& 89.52& 10.48& 28.28\\\hline
\textbf{Non-rollable}&\textbf{$0.273\pm 0.0067$}& 79.62& 91.43& 8.57& 21.38\\\hline
\end{tabular}}
\label{table:Table_2}
\end{table}

As shown, our system could successfully predict graspability affordance. However, it is important to note that the average grasp action error is significantly larger than the average push action error. The mean error was relatively high because the incorrect graspability predictions generated high positional errors that significantly increased the mean error value. This is potentially due to unsuccessful grasps. In the event of an unsuccessful grasp, the object may slowly slip from the robot's hands and land on the table close to its initial position. The object may topple or roll (sometimes off the table), leading to position changes that are uncertain beforehand and therefore cannot be accurately predicted.

\subsection{Planning performance}

\begin{figure}[b]
	\begin{center}
		\includegraphics[width=\columnwidth]{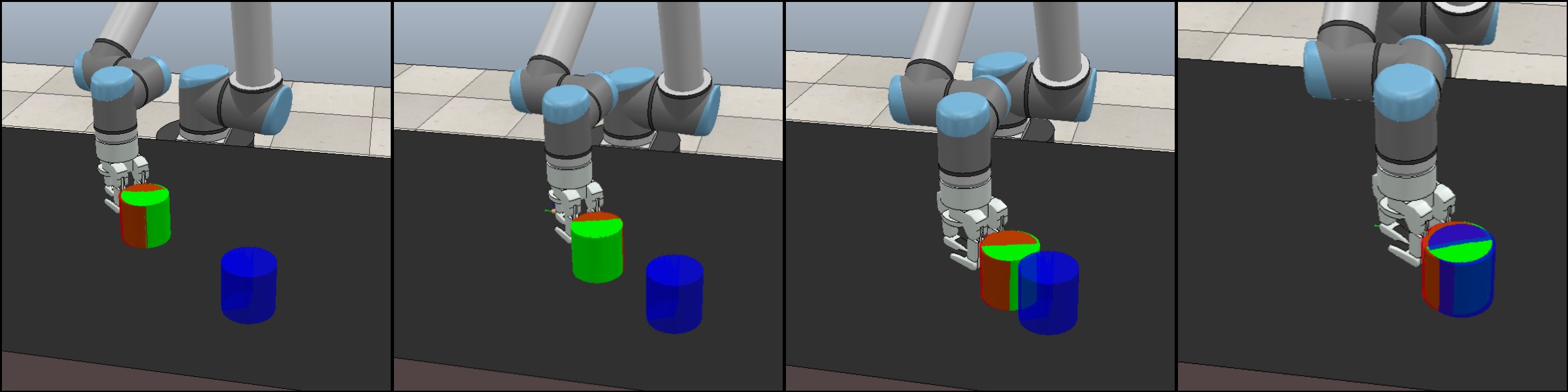}
		\caption{Results of applying the model's predicted actions in a scene. Images are ordered from left to right. The top row is from the first action execution, the bottom row is from the second action execution. The blue object denotes the target position, the red object is generated from the robot's effect predictions. The blue and red objects are not interactable by the robot. The green object is interactable and is acted upon by the robot, to show the ground truth results of the robot's predicted actions.}
		\label{fig:Figure_5}
	\end{center}
\end{figure}

Next, our model is requested to generate plans to bring the object to goal positions that might be beyond the range of single pushes or closer than a full push. Therefore, the planner is expected to generate sequences of actions that might include partial executions as well. A sample plan execution is shown in Figure~\ref{fig:Figure_5}, where the plan is composed of two actions. The goal is shown with a blue box, the actual object positions are shown with red color, and the predictions are shown with green color.

Given goal positions, we run our model in three modes:
\begin{itemize}
    \item predict one push action to reach a goal, which is maximum one step ahead, 
    \item predict one (possibly partial) push action to reach a goal, which is maximum one step ahead,
    \item predict the sequence of (possibly partial) push actions to reach a goal, which is maximum three steps ahead.
\end{itemize}
\hakan{For this part, we experimented with both our model and the LSTM model. However, since the LSTM model does not perform well in long-horizon prediction, in most of our trials, the actions produced by the LSTM model were mostly unusable. During execution, the objects mostly either fell over, slid off the gripper's fingers, or in some cases fell off the table which made the positional error results we obtained unusable for comparison. Because of this, we decided to focus on comparing the planning algorithms instead.  In order to compare their performance we ran a number of trials for these three modes using both our proposed planning method and A* planner.} After plans are made, they are executed by the robot. The distance between the goal position and the final actual position of the object is reported as an error. The results are provided in Table \ref{table:Table_3}. \hakan{For the A* planner,} the object can be brought to the goal position more accurately if the partial actions are considered during planning even in a single-action execution case. Even error obtained in multiple (potentially partial) action executions is smaller than single full action execution. This shows the effectiveness of our system in making accurate plans. \hakan{Across the modes, our proposed method performed better than A* planner with a significant decrease in displacement error. It can also be seen that the error values obtained using our method are very close for all modes.  Since our method can use any segment of the given action primitives for planning, the sampling frequency of the trajectories affects the precision of our system significantly. This effect can be seen in Table \ref{table:Table_3}. By  increasing the sampling frequency, we sampled the same push action paths to trajectories with sample sizes 100 and 300, and this increase in sample size decreased the positional error of the trials significantly. We did not observe any notable difference between the accuracy of  plans made for rollable or non-rollable objects. Unexpected changes during execution like changing the goal position or objects falling over also did not cause any difference in positional error values. } For the results reported in Table~\ref{table:Table_3}; the single full setting is tested with 50 goals, the single partial setting with 100 goals, and the multi partial setting with 80 goals.

\begin{table}
\tbl{The comparison of A* planner and our method for the three experiment modes. The effect of the sampling frequency on our method can be seen from the difference between the mean values of the third and the fourth column. }
{\begin{tabular}{|c|c|c|c|} \hline
\textbf{N-step}&  \textbf{A* Planner}& \textbf{Our Method size=100}& \textbf{Our Method size=300}\\\hline
\textbf{Single Full (m)}&  $0.020 \pm 0.002$& $0.0074 \pm 0.0017$& $0.0021 \pm 0.00044$\\\hline
\textbf{Single Partial (m)}& $0.013 \pm 0.002$& $0.0077 \pm 0.0016$& $0.0022 \pm 0.00068$\\\hline
\textbf{Multi Partial (m)}&  $0.013 \pm 0.002$& $0.0077 \pm 0.0013$& $0.0022\pm 0.00046$\\\hline
\end{tabular}}
\label{table:Table_3}
\end{table}


\section{Conclusion}

In this paper, we realized a model for multi-step action and effect prediction. While previous work's utilization of bidirectional learning is limited, our model specifically creates its latent representations using this concept and is able to make multi-step predictions that are in accordance with ground truth manipulations. We emphasize using object-centric inputs to achieve generalizability and investigate simple affordances of several classes of objects of different sizes. By using a network for single interaction predictions which can be interpreted similarly to a state transition function and pairing it with a planner with heuristics to propose goal-directed actions the model was shown to achieve low error in reaching target positions. While the results of our experiments are promising, the model still requires verification in the real world. Our next step is gathering data and testing our implementation with a real robot. 

Our work uses a conditional architecture to avoid the compounding error problems of recurrent architectures and models that are used in a similar way by feeding their current step output as the next step input. This advantage of using conditional models is shown in this work against using an LSTM network, and against a Multimodal Variational Autoencoder (MVAE) in \cite{seker2022imitation}. Recently, transformer models  gained popularity as being a good alternative to recurrent models. By using the attention mechanism they eliminate the need for recurrence, and attention can potentially be beneficial for our model as well. We plan to investigate the capabilities of such models in the future.

\section{Disclosure statement}

No potential conflict of interest was reported by the authors.

\section*{Acknowledgement(s)}

The numerical calculations reported in this paper were partially performed at TUBITAK ULAKBIM, High Performance and Grid Computing Center (TRUBA resources). The authors would like to thank Alper Ahmetoglu for providing insightful comments for this paper.

\section*{Funding}
This research was supported by TUBITAK (The Scientific and Technological Research Council of Turkey) ARDEB; 1001 program (project number: 120E274); TUBITAK BIDEB; 2210-A program; and by the BAGEP Award of the Science Academy.

\bibliographystyle{tfnlm.bst}
\bibliography{references.bib}

\appendix

\section*{Appendix}

\begin{table}[h]
\tbl{Image Encoder}
{\begin{tabular}{ccc} \hline
\textbf{Layer}&  \textbf{Input Size}& \textbf{Output Size}\\\hline
\textbf{Conv3×3 + LeakyReLU + MaxPool2×2}&  (48, 48, 2)& (24, 24, 32)\\\hline
\textbf{Conv3×3 + LeakyReLU + MaxPool2×2}& (24, 24, 32)& (12, 12, 64)\\\hline
\textbf{Conv3×3 + LeakyReLU + MaxPool2×2}&  (12, 12, 64)& (6, 6, 64)\\\hline
\textbf{Conv3×3 + LeakyReLU + MaxPool2×2}&  (6, 6, 64)& (3, 3, 128)\\\hline
\textbf{Conv3×3 + LeakyReLU + MaxPool2×2}&  (3, 3, 128)& (1, 1, 128)\\\hline
\textbf{Flatten}&  (1, 1, 128)& 128\\\hline
\textbf{Dense}&  128& 16\\\hline
\end{tabular}}
\label{Appendix:0}
\end{table}
\begin{table}[h]
\tbl{Action Encoder}
{\begin{tabular}{ccc} \hline
\textbf{Layer}&  \textbf{Input Size}& \textbf{Output Size}\\\hline
\textbf{Dense + LeakyReLU}&  3& 32\\\hline
\textbf{Dense + LeakyReLU}& 32& 64\\\hline
\textbf{Dense + LeakyReLU}&  64& 64\\\hline
\textbf{Dense + LeakyReLU}&  64& 128\\\hline
\textbf{Dense + LeakyReLU}&  128& 128\\\hline
\textbf{Dense + LeakyReLU}&  128& 256\\\hline
\textbf{Dense + LeakyReLU}&  256& 128\\\hline
\end{tabular}}
\label{Appendix:1}
\end{table}
\begin{table}[h]
\tbl{Effect Encoder}
{\begin{tabular}{ccc} \hline
\textbf{Layer}&  \textbf{Input Size}& \textbf{Output Size}\\\hline
\textbf{Dense + LeakyReLU}&  3& 32\\\hline
\textbf{Dense + LeakyReLU}& 32& 64\\\hline
\textbf{Dense + LeakyReLU}&  64& 64\\\hline
\textbf{Dense + LeakyReLU}&  64& 128\\\hline
\textbf{Dense + LeakyReLU}&  128& 128\\\hline
\textbf{Dense + LeakyReLU}&  128& 256\\\hline
\textbf{Dense}&  256& 128\\\hline
\end{tabular}}
\label{Appendix:2}
\end{table}
\begin{table}
\tbl{Action Decoder}
{\begin{tabular}{ccc} \hline
\textbf{Layer}&  \textbf{Input Size}& \textbf{Output Size}\\\hline
\textbf{Dense + LeakyReLU}&  1024& 512\\\hline
\textbf{Dense + LeakyReLU}& 512& 256\\\hline
\textbf{Dense + LeakyReLU}&  256& 128\\\hline
\textbf{Dense + LeakyReLU}&  128& 32\\\hline
\textbf{Dense + LeakyReLU}&  32& 4\\\hline
\end{tabular}}
\label{Appendix:3}
\end{table}
\begin{table}[h]
\tbl{Effect Decoder}
{\begin{tabular}{ccc} \hline
\textbf{Layer}&  \textbf{Input Size}& \textbf{Output Size}\\\hline
\textbf{Dense + LeakyReLU}&  1024& 512\\\hline
\textbf{Dense + LeakyReLU}& 512& 256\\\hline
\textbf{Dense + LeakyReLU}&  256& 128\\\hline
\textbf{Dense + LeakyReLU}&  128& 32\\\hline
\textbf{Dense + LeakyReLU}&  32& 4\\\hline
\end{tabular}}
\label{Appendix:4}
\end{table}

\begin{table}[h]
\tbl{Overall structure of the model}
{\begin{tabular}{ccc} \hline
\textbf{Layer}&  \textbf{Input Size}& \textbf{Output Size}\\\hline
\textbf{Object Encoder}&  (48, 48, 2)& 16\\\hline
\textbf{Action Encoder}& 3& 128\\\hline
\textbf{Effect Encoder}&  3& 128\\\hline
\textbf{Concatanate - Action + Object + $t_q$ }&  128 + 16 + 1& 145\\\hline
\textbf{Concatanate - Effect + Object + $t_q$ }&  128 + 16 + 1& 145\\\hline
\textbf{1D Avarage - Effect + Action}&  145 + 145& 145\\\hline
\textbf{Dense}&  145& 1024\\\hline
\textbf{Action Decoder}&  1024& 4\\\hline
\textbf{Effect Decoder}&  1024& 4\\\hline
\end{tabular}}
\label{Appendix:5}
\end{table}

\begin{algorithm}[h]
\caption{Proposed Planning Algorithm (without online checking)}\label{alg:cap}
\begin{algorithmic}
\State $actionSequence \gets [\;]$
\State $successValue \gets $ \Comment{Can be set to any success distance}
\While{Object is not in the desired position}
\State  $direction \gets sampleRandomPointOnCircle()$ 
\While{$direction$ changes} \Comment{won't change after solution's been found }
\State $ predictedAction, predictedEffect \gets model.predict([t=1,direction])$
\State $loss \gets meanSquaredError(targetPosition, predictedEffect)$
\State $ gradient \gets calculateGradient(direction,loss)$
\State $applyGradient(gradient,direction)$
\EndWhile
\State $fullActionTraj,fullEffectTraj \gets model.predict([t,direction]),0 \leq t \leq 1$ 
\State $index \gets 0$
\State $done \gets false$
\While{$index < size(fullEffectTraj)$}
\If{$successValue > distanceBetween(target,fullEffectTraj[index])$}
    \State $actionSequence.append(fullActionTraj[0:index])$
    \State $done \gets true$
    \State $break$
\EndIf
\EndWhile
\If{$done=false$}
    \State $actionSequence.append(fullActionTraj)$
\EndIf
\If{$done=true$}
\State $break$
\EndIf
\EndWhile
\end{algorithmic}
\end{algorithm}


\end{document}